\documentclass[runningheads]{llncs}
\usepackage{graphicx}
\usepackage{mathalfa}
\DeclareMathAlphabet{\mathpzc}{OT1}{pzc}{m}{it}
\usepackage{amsfonts}
\usepackage{amsmath}
\DeclareMathOperator*{\argmax}{argmax}
\usepackage{algorithm}
\usepackage{algpseudocode}
\usepackage{amssymb}
\usepackage{caption} 
\begin{document}
\title{Deep Reinforcement Learning for Resource Allocation in Business Processes}
%
%
\renewcommand{\lastandname}{\unskip,}

\author{Kamil Żbikowski\and
Michał Ostapowicz\and
Piotr Gawrysiak}

\authorrunning{K. Żbikowski et al.}
%
\institute{Warsaw University of Technology, ul. Nowowiejska 15/19, 00-665 Warsaw, Poland
\email{kamil.zbikowski@pw.edu.pl},\email{michal.ostapowicz@pw.edu.pl},\email{piotr.gawrysiak@pw.edu.pl}}
\maketitle              
\begin{abstract}
Assigning resources in business processes execution is a repetitive task that can be effectively automated. However, different automation methods may give varying results that may not be optimal. Proper resource allocation is crucial as it may lead to significant cost reductions or increased effectiveness that results in increased revenues. 

In this work, we first propose a novel representation that allows modeling of a multi-process environment with different process-based rewards. These processes can share resources that differ in their eligibility. Then, we use double deep reinforcement learning to look for optimal resource allocation policy. We compare those results with two popular strategies that are widely used in the industry. Learning optimal policy through reinforcement learning requires frequent interactions with the environment, so we also designed and developed a simulation engine that can mimic real-world processes. 

The results obtained are promising. Deep reinforcement learning based resource allocation achieved significantly better results compared to two commonly used techniques. 

\end{abstract}
\section{Introduction}
In process science there is a wide range of approaches that are employed in different stages of operational processes' life cycles. Following \cite{van2016data}, these include, among others, optimization and stochastic techniques. Business processes can be also categorized according to the following perspectives: control-flow, organizational, data and time perspective \cite{van2013business}. Resource allocation is focused on the organizational perspective utilizing optimization and stochastic approaches. 

As it was emphasised in \cite{huang2011mining} resource allocation, while being important from the perspective of processes improvement, did not receive much attention at the time. However, as it was demonstrated in \cite{arias2018human} the problem received much more attention in the last decade, what was reflected in the number of published scientific papers.

This paper addresses the problem of resource allocation with the use of methods known as approximate reinforcement learning. We specifically applied recent advancements in deep reinforcement learning such as double deep q-networks (double DQN) described in \cite{mnih2015human}. To use those methods we firstly propose a representation of business processes suite that helps to design architecture of neural networks in terms of appropriate inputs and outputs. 

To the best of our knowledge this is the first work that proposes a method utilizing a double deep reinforcement learning for an on-line resource allocation for multiple-process and multi-resource environment. Previous approaches either used so called "post mortem" data in the form of event logs (e.g. \cite{liu2012mining}), or applied on-line learning, but due to the usage of tabular algorithms were limited by the exploding computational complexity when the number of possible states increased.

In the next section we provide an overview of reinforcement learning methods and outline improvements of deep learning approaches over existing solutions. Then we analyze and discuss different approaches to resources allocation. In Section \ref{sec:app} we outline our approach for modelling operational processes for the purpose of training resource allocation agents. In Section \ref{sec:evaluation} we describe the simulation engine used in training and its experimental setup. In Section \ref{sec:res} we evaluate the proposed approach and present outcomes of the experiments. In Section \ref{sec:conc} we summarize the results and sketch potential future research directions. 

\section{Background and Related Work}
\subsection{Deep Reinforcement Learning} 
Following \cite{sutton2018reinforcement}, reinforcement learning is "learning what to do -- how to map situations to actions -- so as to maximize a numerical reward signal". There are two main branches of reinforcement learning, namely tabular and approximate methods. The former provide a consistent theoretical framework that under certain conditions guarantees convergence. Their disadvantage is increasing computational complexity and memory requirements when the number of states grows. The latter are able to generalise over large number of states but do not provide any guarantee of convergence. 

The methods that we use in this work find optimal actions indirectly, identifying  optimal action values for each state-action pair. Following recursive Bellman equation for the state-action pair \cite{sutton2018reinforcement}:
\begin{equation}
    \label{eq:bellman}
    q_{\pi}(s,a) = \sum_{s',r} p(s',r|s,a)[r + \gamma \sum_{a'}\pi(a'|s')q_{\pi}(s',a')]
\end{equation}
an optimal policy is a policy that at each subsequent step takes an action that maximizes state-action value, that is $q_{*}(s,a) = max_{\pi} q_{\pi}(s,a)$. In the above equation $p(s',r|s,a)$ is a conditional probability of moving to state $s'$ and receiving reward $r$ after taking action $a$ in state $s$; $\pi_(a|s)$ is the probability of taking action $a$ in state $s$; $\gamma \in [0,1]$ is a discount factor.

When we analyse equation \ref{eq:bellman} we can intuitively understand problems with iterative tabular methods for finding optimal policy $\pi^*$ for high-dimensional state spaces. Fortunately, recent advancements in deep learning methods allow to further enhance approximate reinforcement learning methods with a most visible example being human level results for Atari suite \cite{bellemare2013arcade} obtained with the use of double deep Q-network \cite{mnih2013playing}.

\subsection{Resource Allocation}
In \cite{arias2018human} we can find a survey of human resource allocation methods. The spectrum of approaches is wide. In \cite{huang2012resource}, \cite{zhao2015optimization}, \cite{arias2016framework}, \cite{havur2016resource} and \cite{xu2008resource} we can find solutions based on static, rule based algorithms.

There are number of approaches for resource allocation that rely on applying predictive models. In \cite{park2019prediction} an offline prediction model based on LSTM is combined with extended minimum cost and maximum flow algorithms. 

In \cite{huang2011reinforcement} authors introduce Reinforcement Learning Based Resource Allocation Mechanism that utilizes Q-learning for the purpose of resource allocation. For handling multiple business processes the queuing mechanism is applied.

Reinforcement learning has been also used for the task of proactive business process adaptation \cite{metzger2020triggering} \cite{huang2010adaptive}. The goal there is to monitor the particular business process case while it is running and intervene in case of any detected upcoming problems. 

The evaluations conducted in aforementioned works are either based on simulations (\cite{huang2010adaptive}, \cite{huang2011reinforcement}) or on analysis of historical data, mostly from Business Process Intelligence Challenge (\cite{metzger2020triggering}, \cite{park2019prediction}, \cite{palm2020online}). The latter has an obvious advantage of being real-world based dataset while at the same time being limited by the number of available cases. The former offers potentially infinite number of cases, but alignment between simulated data and real business processes is hard to achieve.

In \cite{silvander2019business} authors proposed a deep reinforcement learning method for business process optimization, however their research objective is concentrated on analysing which parameters of DQN are optimal. 

\section{Approach}
\label{sec:app}
This section describes the methods that we used to conduct the experiment. First, we will introduce concepts related to business process resource allocation. Then we will present double deep reinforcement learning \cite{van2016deep}for finding optimal resource allocation policy.

As it was pointed earlier, both tabular and approximate algorithms in the area of reinforcement learning require frequent interaction with the execution environment. For the purpose of this work, we designed and developed a dedicated simulation environment that we call Simulation Engine. However, it can serve as well as a general-purpose framework for testing resource allocation algorithms. Concepts that we use for defining business process environment assume existence of such an engine. They incorporate parameters describing the level of uncertainty in regard to their instances. The purpose here is to replicate stochastic behavior during process execution in the real-world scenarios. 

We imagine a business process workflow as a sequence of tasks\footnote{Task here should not be confused with the task definition used in reinforcement learning literature where it actually means the objective of the whole learning process. In the RL sense, our task would be to "solve" Business Process Suite (meaning obtaining as much of cumulative reward as possible) in the form of Definition \ref{def:bpe}.} that are drawn from the queue and are being executed by adequate resources (both human and non-human). Each task realization is in fact an instance of a task specification described below. Task here is considered as an unbreakable unit of work that a resource can be assigned to and works on for a specified amount of time. 

\begin{figure}
    \includegraphics[scale=0.5]{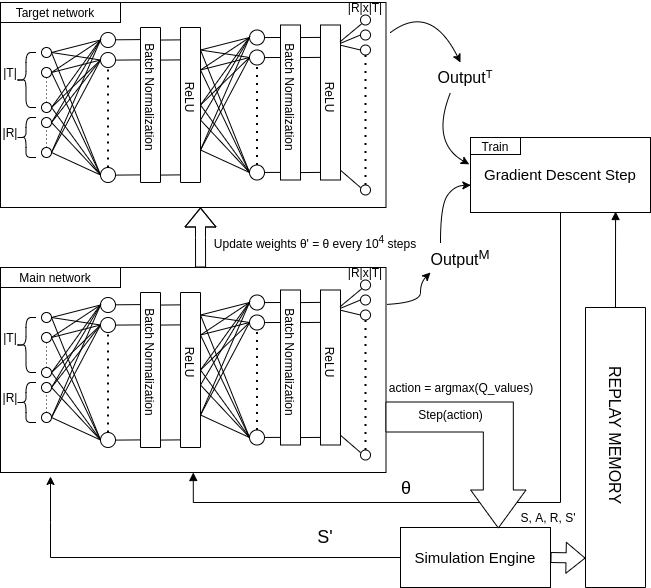}
    \centering
    \caption{Training architecture diagram. Learning process is centered around Simulation Engine that takes action from the main network and returns the reward and the next state.} 
    \label{fig:arch}
\end{figure}

\begin{definition}[Task]
Let the tuple $\mathpzc{(i, C^i, d, s, b)}$ define a task $\mathpzc{t_i}$ that is a single work unit represented in the business process environment where:
\begin{itemize}
    \item $\mathpzc{i}$ is a unique task identifier where $\mathpzc{i} \in \{0,1,2,...\}$,
    \item $\mathpzc{C^i}$ is a set of transitions from a given task $i$,
    \item $\mathpzc{d} \in \mathbb{R^+}$ is a mean task duration with $\mathpzc{s}$ being its standard deviation and
    \item $\mathpzc{b} \in \{0,1\}$ indicates whether it is a starting task for particular business process.
\end{itemize}
\end{definition}

Each task in the business process (see e.g. Figure \ref{fig:bp1}) may have zero or more connections from itself to other tasks.

\begin{definition}[Task Transition]
For a given task $\mathpzc{t_i}$ a task transition $\mathpzc{c^i_j}$ is a tuple $\mathpzc{(j, p)}$ where $\mathpzc{j}$ is an unique identifier of a task that this transitions refers to where $\mathpzc{p}$ is a probability of this transition. If $\mathpzc{i=j}$ it is a transition to itself.
\end{definition}

\begin{definition}[Resource]
Let the tuple $\mathpzc{(k)}$ define a single resource $\mathpzc{r_k}$ where $\mathpzc{k} \in \{0,1,2,...\}$ is a unique resources identifier. To refer to the set of all resources we use $\mathpzc{\hat{R}}$.
\end{definition}

\begin{definition}[Resource Eligibility]
If a resource $\mathpzc{r_k}$ can be assign to a task $\mathpzc{t_i}$ it is said it is eligible for this task. Set $\mathpzc{E^i} = \{\mathpzc{e_k^i} : \mathpzc{e_k^i} \in R^+\}$ contains all resource eligibility modifiers for a given task $\mathpzc{i}$. The lower the $\mathpzc{e_k^i}$, the shorter is the expected execution of task $\mathpzc{t_i}$. To refer to the set of all properties of eligibility for all defined resources $\mathpzc{\hat{R}}$ we use $\mathpzc{\hat{E}}$.
\end{definition}

\begin{definition}[Business Process]
\label{def:bp}
Let a tuple $\mathpzc{(m, f_m,}\mathbb{R}_\mathpzc{m},\mathpzc{T_m)}$ define a business process $\mathpzc{P_m}$ where $\mathpzc{m}$ is an unique identifier of a process $\mathpzc{P_m}$ and $\mathpzc{T_m}$ is a set of tasks belonging to the process $\mathpzc{P_m}$ and $\mathpzc{ t_i \in T_m \implies \neg \exists n : n \neq m \land t_i \in T_n}$. Relative frequency of a particular business process is defined by $\mathpzc{f_m}$. By $\mathbb{R}_\mathpzc{m}$ we refer to the reward that is received by finishing this business process instance. To refer to the set of all defined business processes we use $\mathpzc{\hat{P}}$.
\end{definition}

An example of a business process can be found in Figure \ref{fig:bp1}. Nodes represent tasks and their identifiers. Arrows define possible task transitions from particular nodes. The numbers on arrows represent transition probabilities to other tasks.

\begin{definition}[Business Process Suite]
\label{def:bpe}
Let a tuple $(\mathpzc{\hat{R}}, \mathpzc{\hat{E}}, \mathpzc{\hat{P}})$ define a Business Process Suite that consists of a resources set $\mathpzc{\hat{R}}$, resources eligibility set $\mathpzc{\hat{E}}$ and business processes set $\mathpzc{\hat{P}}$ such that: $\forall \mathpzc{r_k} \in \mathpzc{\hat{R}} \; \exists m,i \; \mathpzc{e_k^i} \in \mathpzc{\hat{E}} \land \mathpzc{t_i} \in  \mathpzc{T_m} \land \mathpzc{T_m} \in \mathpzc{\hat{P}}$ and $\nexists \mathpzc{r_k \in \hat{R}},i,m \; \mathpzc{e_k^i} \in \mathpzc{\hat{E}} \land \mathpzc{t_i} \notin  \mathpzc{T_m} \land \mathpzc{T_m} \in \mathpzc{\hat{P}}$
\end{definition}

Business Process Suite is a meta definition of the whole business processes execution environment that consist of tasks that aggregate to business processes and resources that can execute tasks in accordance with the defined eligibility. We will refer to the instances of business processes as business process cases. 

\begin{definition}[Business Process Case]
\label{def:bpc}
Let a tuple $(\mathpzc{P_m, i, o})$  define a business process case $\mathpzc{\tilde{P_m}}$ where $\mathpzc{P_m}$ is a business process definition, $\mathpzc{i}$ is a current task that is being executed and $\mathpzc{o} \in \{0,1\}$ is information whether it is running (0) or was completed (1). 
\end{definition}

\begin{definition}[Task Instance]
\label{def:ti}
Let a tuple $(\mathpzc{i, r_k})$ be a task instance $\mathpzc{\tilde{t_i}}$. At a particular moment of execution there exists exactly one task instance matching business process case property $\mathpzc{i}$. The exact duration is determined by properties $\mathpzc{d}$ and $\mathpzc{s}$ of task definition $\mathpzc{t_i}$.
\end{definition}

\begin{definition}[Task Queue]
\label{def:tq}
Let the ordered list $\mathpzc{(N^{t_0}, N^{t_1}, N^{t_2}, ..., N^{t_i})}$ define a task queue that stores information about number of task instances $\mathpzc{N^{t_i}}$ for a given task $\mathpzc{t_i}$.
\end{definition}

\begin{property}
Direct consequence of definitions \ref{def:bp}, \ref{def:bpc}, \ref{def:ti} and \ref{def:tq} is that number of task instances in the task queue matching the definition of task with identifiers from particular business processes is equal to the number of business process cases.
\end{property}

The process of learning follows schema defined in \cite{mnih2013playing} and \cite{mnih2015human}. We use two sets of weights $\theta$ and $\theta'$. The former is used for the online learning with random mini-batches sampled from a dedicated experience replay queue $\mathpzc{D}$. The latter is updated periodically to the weights of the more frequently changing counterpart. The update period used in tests was $10^4$ steps. Detailed algorithm is outlined in listing \ref{list:a1}. 

\begin{algorithm}
  \caption{Double DQN training loop}\label{euclid}
  \label{list:a1}
  \begin{algorithmic}[1]
    \State Initialize number of episodes $E$, and number of steps in episode $M$
    \State Initialize batch size $\beta$ \Comment{Set to 32 in tests}
    \State Initialize randomly two sets of neural network weights $\theta$ and $\theta'$
    \State $\mathpzc{D} := \{\}$ \Comment{Replay memory of size $E * M * 0.1$}
    \State Initialize environment $\mathpzc{E}$
    \For{e=0 \textbf{in} E}
        \State S := \Call{Reset}{$\mathpzc{E}$}
        \For{m=0 \textbf{in} M}
            \If{\Call{Random}  $< \epsilon$}
                \State a := \Call{SelectRandomAction}{}
            \Else
                \State a := $argmax_a Q(S, a; \theta)$
            \EndIf
            \State $\mathbb{S'}$, $\mathbb{R}$ := \Call{Step}{$\mathpzc{E}$, a}
            \State Put a tuple $(S, a, R, S')$ in $\mathpzc{D}$
            \State Sample $\beta$ experiences from $\mathpzc{D}$ to $\mathbb{(S, A, R, S')}$
            \State $Q_{target} := \mathbb{R} + \delta *  Q(\mathbb{S'}, \argmax_a Q(\mathbb{S'},a;\theta);\theta')$
            \State $Q_{current} := Q(S,a;\theta)$
            \State $\theta_{t+1} = \theta_{t} + \nabla_{\theta_t} (Q_{target} - 
            Q_{current})^2 $
            \State Each $10^4$ steps update $\theta' := \theta$
        \EndFor
    \EndFor
  \end{algorithmic}
\end{algorithm}

In RL there exists separation between continuing and episodic RL tasks \cite{sutton2018reinforcement}. The former are ending in a terminal state and differ in the rewards for the different outcomes. The latter are running infinitely and accumulate rewards over time. Business processes suite is a continuing RL task in its nature. However, in our work we artificially terminate each execution after $M$ steps simulating episodic environment. We observed that it gave much better results than treating the whole set of business processes as a continuing learning task. As it is shown in Section \ref{sec:evaluation} agents trained in such a way can be used in a continuing setup without loss of their performance. 

\section{Evaluation}
\label{sec:evaluation}
This section describes the setup of the experiments that we have conducted to assess proposed methods. Firstly, we briefly outline the data flow of a learning process described in Algorithm \ref{list:a1}. Secondly, we present parametrization of a business process suite used for the evaluation.

In the Figure \ref{fig:arch} an architecture of a system used in the experiment is presented in accordance with main data flows. It is a direct implementation of the training algorithm described in Algorithm \ref{list:a1}. We used two neural networks: main and target. Both had the same architecture consisting of one input layer with $\mathpzc{|R| + |T|}$ inputs, two densely connected hidden layers  containing $32$ neurons each and one output layer with $\mathpzc{|R| x |T|}$ outputs. After each hidden layer there is Batch Normalization layer \cite{ioffe2015batch}. Its purpose is to scale each output from hidden neuron layer before computing activation function. This operation improves training speed by reducing the undesired effects such as vanishing / exploding gradient updates. 

In terms of inputs we tested several configurations. One that gave the most promising results is defined as follows: 
\begin{equation}
    \mathbb{S} = [\rho_0, \rho_1,...\rho_{\mathpzc{|R|-1}}, \zeta_0, \zeta_1, ..., \zeta_{|\mathpzc{T}|-1}]
    \label{eq:st}
\end{equation}
where $\rho_{\mathpzc{k}} = \mathpzc{i}$ refers to particular resource assignment to one of its eligible task, and $\zeta_{\mathpzc{i}} = \mathpzc{N}^{t_i} / \sum_{l=0}^\mathpzc{|T|-1} \mathpzc{N}^{t_l}$ is a relative load of a a given task to all the tasks present in the task queue. 

Outputs of the neural network are an approximation of a q-value for each of the available actions. An action here is assigning a particular resource to a particular task or taking no action for a current time-step. Thus, number of outputs equals $\mathpzc{|R|}\mathpzc{|T|} + 1$. This number grows quickly with the number of resources and tasks. This in turn may lead to the significant increase in training time or even inability to obtain adequate q-value estimation.  

It is worth mentioning that our objective was not to incorporate any prior knowledge into the model. An exemplification of this approach is ignoring the information about resources' eligibility. This seemingly strict constraint makes potential applications a lot broader as inferring eligibility of resources is not always straightforward in the real-world scenarios. 

\subsection{Simulation Engine}

Simulation engine works in a similar manner to Atari environments described in \cite{mnih2013playing}. It takes an action from the agent and executes it on the business process suite changing its state. 

Simulation Engine is based on two sets of tasks' instances: enabled and current tasks. The former consists of tasks waiting to be executed, without allocated resources. The latter consists of tasks that are already being executed. Enabled tasks set has a limit of total duration of tasks in it which can be modified via one of the configuration files. It prevents the enabled tasks set from expanding uncontrollably. 

The main part of the simulation engine is step() method that is called in every iteration of the training loop.  At the beginning of step(), according to the simulation configuration, a new process case may appear. It goes straight to the enabled tasks set. The only parameter passed to step() is a two-element tuple made up of task id and resource supposed to be allocated to that task. If there is a task with the task id given in action in the enabled task set and an action resource is eligible for that task, that particular resource is allocated.

The allocation process begins with removing the resource from the list of available resources and also removing the task from the enabled tasks set. In the next move, the task duration is calculated and that task is added to the current tasks set. If this task has no successors, it is the last task of the business process case and the step() method will return the reward for completing this business process case. In the next step, regardless of whether the resource was successfully allocated or not, the engine iterates through the current tasks set and decrements their duration. If the duration of the task is equal to 0, it is removed from the current task set. Then, the resource that handled it is added back to the list of available resources. Depending on the transition probabilities, the successor of the completed task is selected and added to the enabled task set. The engine also allows restoring execution environment to its basic state.

Simulation engine can be configured to work with following state representations: 

\begin{itemize}
    \item \textbf{std} - state is a $\mathpzc{|R| x |T|}$ matrix. Each matrix element can be one of three possible values -1, 0, and 1. The value of 1 represents a situation when a particular resource is allocated to the particular task, 0 when a resource is eligible for a given task and that task is in enabled tasks set and -1 otherwise. 
    \item \textbf{a1} - state is a vector of size $\mathpzc{|R| + |T|}$, where the first $\mathpzc{|R|}$ elements shows how the resources are allocated. The value can be either id of the task to which resource is allocated or -1 if the resource is not allocated at all. Each element of the second part (of size $\mathpzc{|T|}$) equals to number of tasks present in enabled tasks set.
    \item \textbf{a10} - state representation is identical to \textbf{a1}, but instead of the number of tasks it stores the percentage of all task calculated in regards to the size of the whole enabled task set.
    \item \textbf{a2} - extends \textbf{a1} with resource eligibilities vector (of size $\mathpzc{|R| x |T|}$) with 1 if the resource is eligible to the task and -1 otherwise.
\end{itemize}

In our experiments, we used the \textbf{a10} representation which is defined in Equation \ref{eq:st}.

\subsection{Experimental setup}
To evaluate proposed method we devised a business processes suite containing two business processes $\mathpzc{m}=0$ and $\mathpzc{m}=1$. Although they are quite small in terms of number of tasks, the tasks transitions are non deterministic which intuitively makes the learning process harder. 

\begin{figure}
    \includegraphics[scale=0.5]{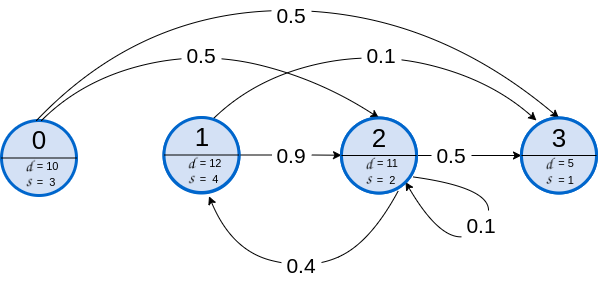}
    \centering
    \caption{Graph of a first business process used in evaluation.} \label{fig:bp1}
\end{figure}

\begin{figure}
    \includegraphics[scale=0.5]{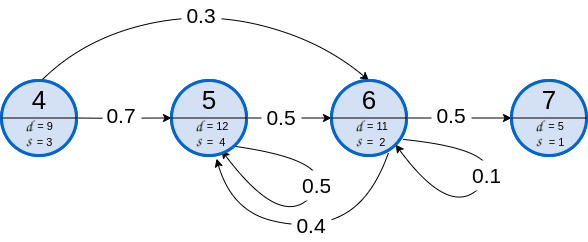}
    \centering
    \caption{Graph of a second business process used in evaluation.} \label{fig:bp2}
\end{figure}

In Figures \ref{fig:bp1} and \ref{fig:bp2} we can see both processes' graphs along with information about their tasks' parametrization. In Table \ref{tab:res_ef} we can see available resources from the testing suite along with the information about their eligibility in regard to particular tasks. 

Both processes have same reward $\mathbb{R}_\mathpzc{0}=\mathbb{R}_\mathpzc{1} = 1$ which is received for each completed business process case. They differ in their relative frequency, which for first process is $\mathpzc{f}_0 = 1$ and $\mathpzc{f}_1 = 6$ for the second one. 

\begin{table}
	\centering
	\renewcommand{\arraystretch}{1.1}
	\setlength{\tabcolsep}{2em}
	\begin{tabular}{cccc}
        \hline
		&\multicolumn{3}{c}{Resources} \\ \cline{2-4}
		Task ID&0&1&2 \\
		\hline
		0  & -&0.75&2.8 \\
		 1  &1.4&	0.3&	- \\
		 2  &0.3&	-&	2.7 \\
		 3  &-&	2.7&	0.1 \\
		 4  &0.6&	2.6&	- \\
		 5  &0.4&	-&	10.5 \\
		 6  &1.1&	-&	1.7 \\
		 7  &0.4&	0.6&	2.5 \\
	\end{tabular}     
	\caption{Resource eligibility. Values in cells define resource efficiency that is used in Simulation Engine. Final duration is obtained by multiplying duration $\mathpzc{d}$ of a particular task by the adequate value form the table. Lack of value indicates that particular resource is not eligible for a given task. }
	\label{tab:res_ef}
\end{table}

In terms of an algorithm parametrization, we set number of episodes $E$ to $600$ and number of steps in a single episode to $400$. $\epsilon$ according to \cite{mnih2015human} was linearly annealed from $1$ to $0.1$ over first $E*M*0.1$ steps. Size of the memory buffer was set $E*M*0.1$ elements.

\section{Results and discussion}
\label{sec:res}
\begin{figure}
    \includegraphics[width=\textwidth]{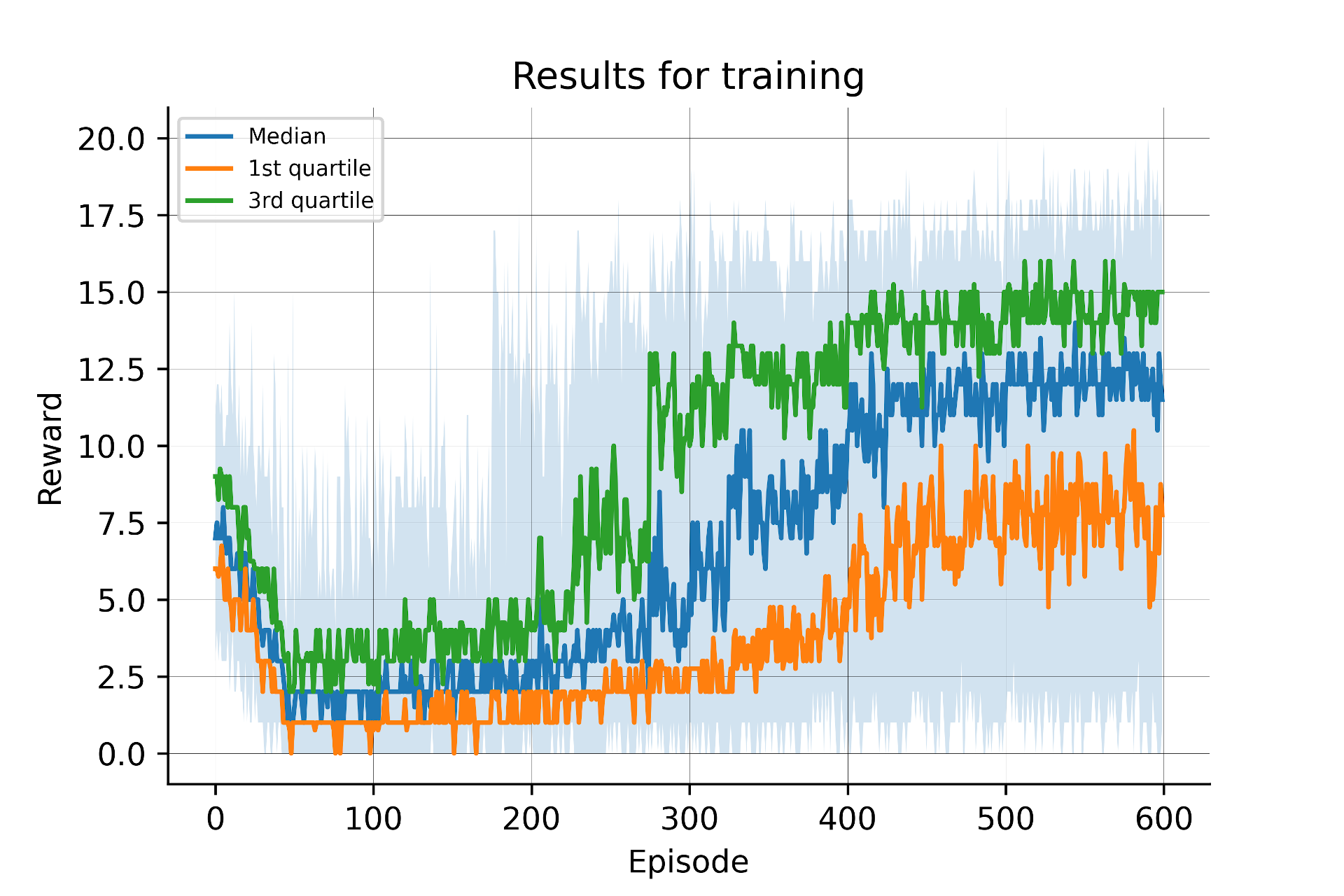}
    \caption{Sum of rewards for subsequent episodes during training on the test suite over 30 training runs.}
    \label{fig:train}
\end{figure}

We run $30$ tests for the test suite. The results are presented in the Figure \ref{fig:train}. We can see that the variance in the cumulative sum of rewards is tremendous. Best models achieve up to $20$ units of reward while the worst keep their score around zero. 

Our findings are consistent with the general perception of how deep reinforcement learning works \cite{rlblogpost}. In particular, training model that achieves satisfactory results strongly depends on weights initialization. 

\begin{figure}
    \includegraphics[width=\textwidth]{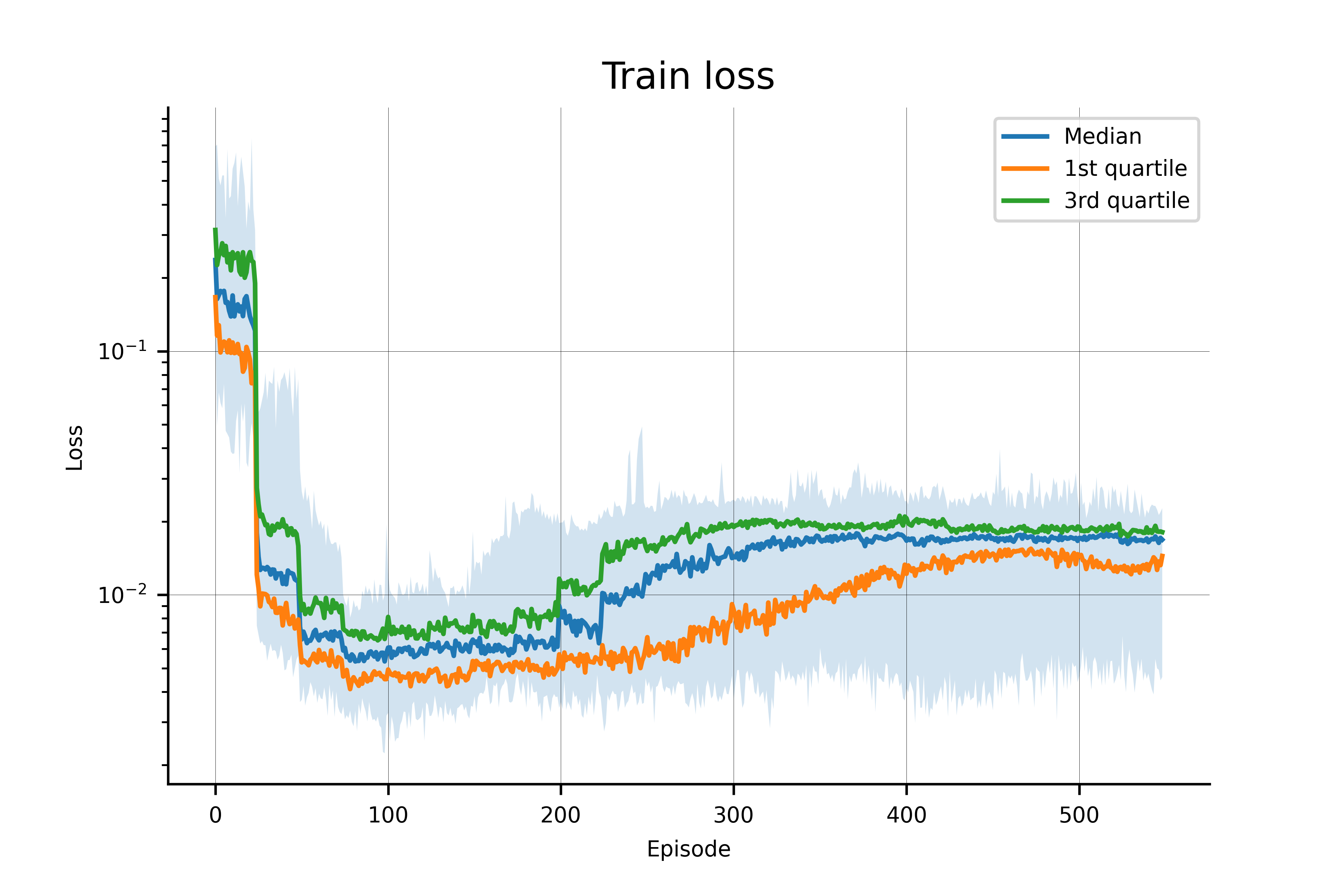}
    \caption{Training loss between target and main networks.}
    \label{fig:train_loss}
\end{figure}

As we can see in Figure \ref{fig:train_loss} the value of a loss function also varies significantly. Moreover, its value after initial drop steadily increases with subsequent episodes. This is a phenomena that is characteristic for DQN. The error measures the difference of training and main network outputs. This value is not directly connected with optimization target - maximizing the cumulative reward over all steps. 

\begin{figure}
    \includegraphics[width=\textwidth]{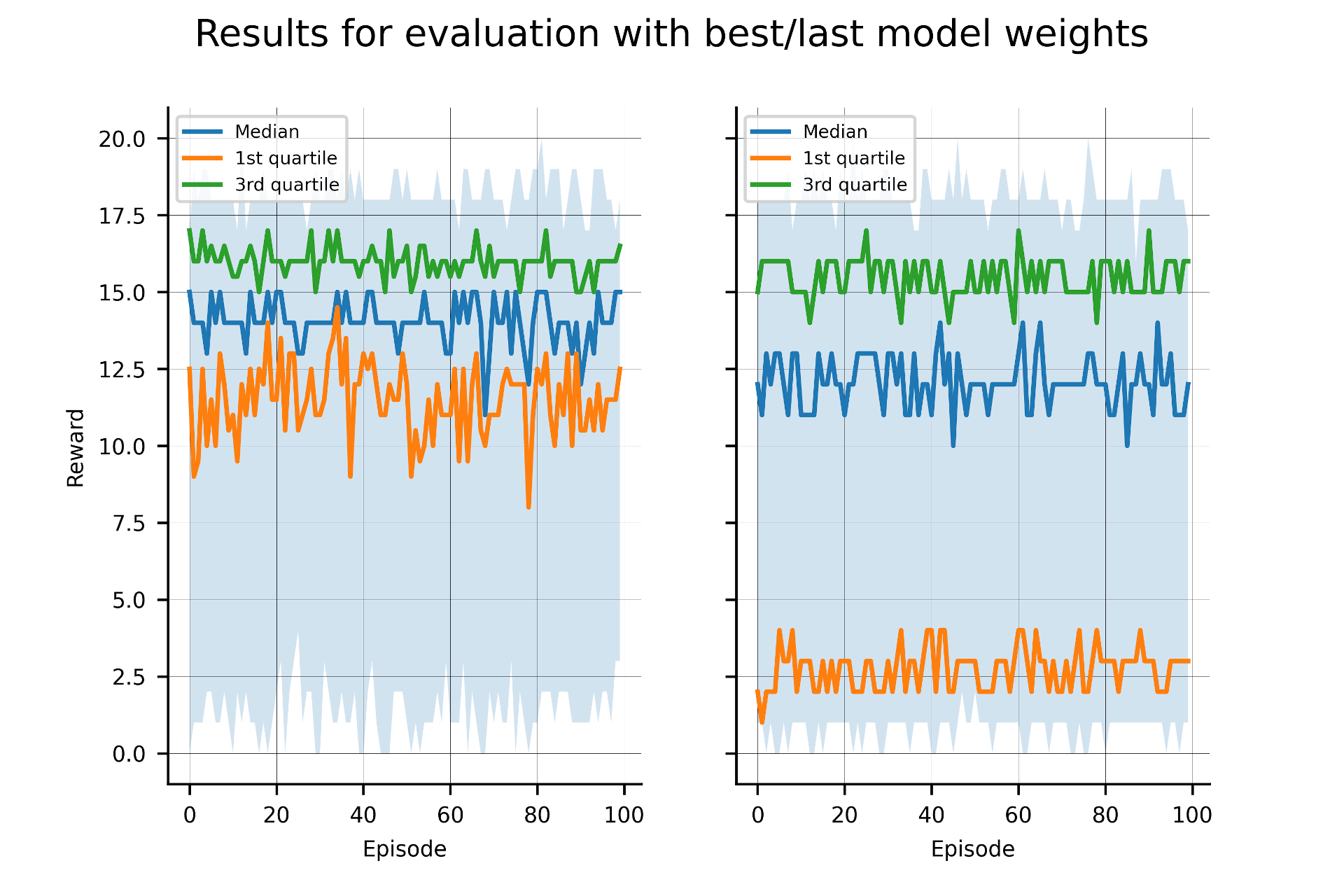}
    \caption{Results obtained by best (left) and last (right) models over 30 runs. } \label{fig:test}
\end{figure}

In \cite{mnih2015human} authors recommend saving model parameters if they are better than the best previously seen (in terms of cumulative reward) during the current training run. This approach allows to address - to some extent - a catastrophic forgetting effect and overall instability of approximate methods. For each run we save both best and last episode's weights. After the training phase we got $30$ models as a result of keeping parameters giving highest rewards during learning and $30$ models with parameters obtained at the end of training. The distribution over all runs can be seen in Figure \ref{fig:test}. We can see that the models with best parameters achieve significantly higher cumulative rewards. Median averaged over $100$ episodes was $14.04$ for the best set of parameters and $12.07$ for the last set. 

To asses the results obtained by the deep learning agent we implemented two commonly used heuristics:

\begin{itemize}
    \item FIFO (first in, first out) - the first-in first-out policy was implemented in an attempt to avoid any potential bias while resolving conflicts in resource allocation. In our case, instead of considering task instances themselves, we try to allocate resources to the business process cases that arrived the earliest.
    \item SPT (shortest processing time) - our implementation of shortest processing time algorithm tries to allocate resources to the task instances that take the shortest time to complete (without taking into account resource efficiencies for tasks). Thanks to this policy, we are able to prevent the longest tasks from occupying resources when these resources could be used to complete other, much shorter tasks and therefore shorten the task queue. 
\end{itemize}

\begin{figure}
    \includegraphics[width=\textwidth]{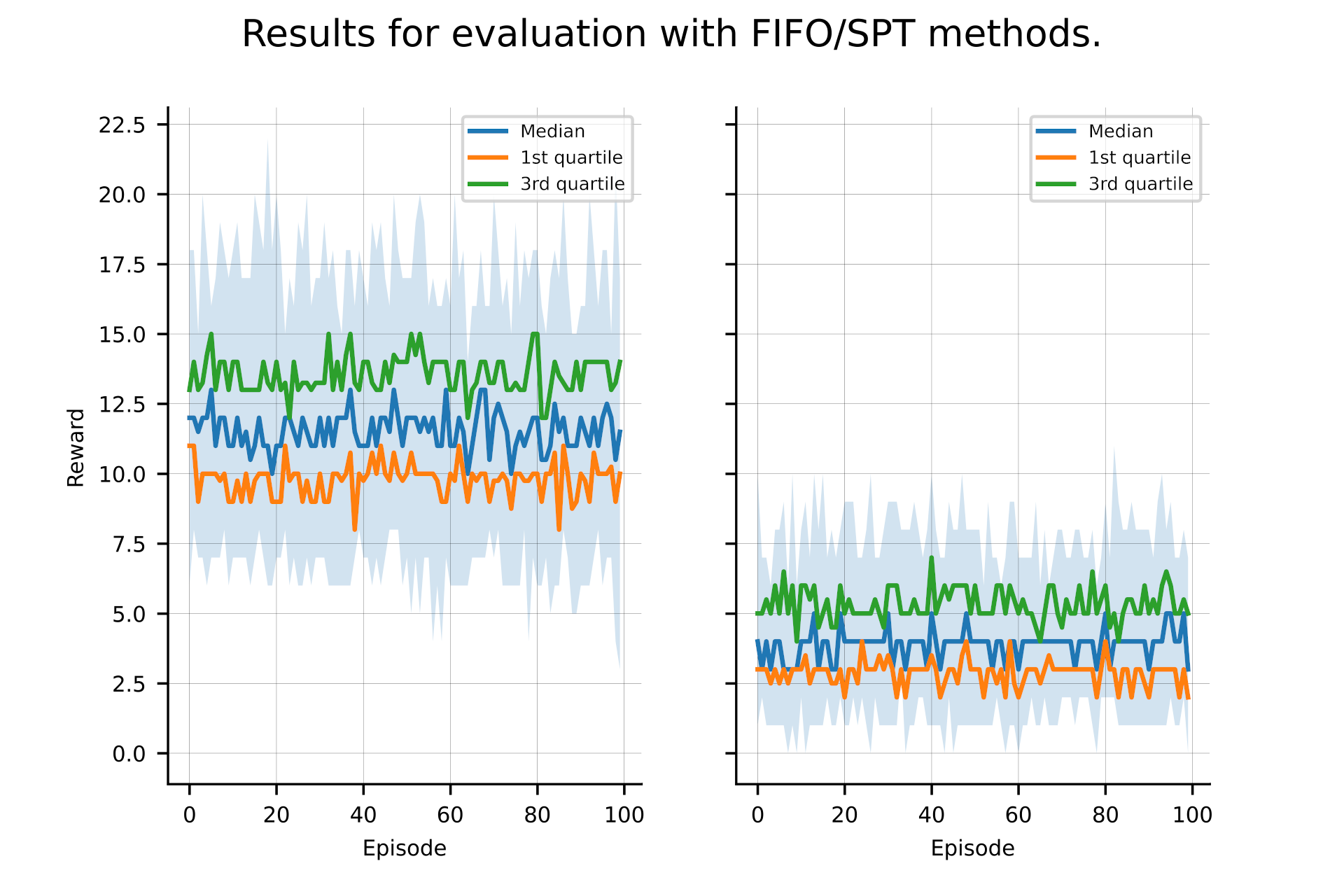}
    \caption{Results obtained for FIFO and SPT algorithms over 30 runs. } \label{fig:test_comp}
\end{figure}

We conducted the same test lasting $100$ episodes for both heuristics. Results are presented in Figure \ref{fig:test_comp}. Median averaged  over $100$ episodes was $11.54$ for FIFO and $3.88$ for SPT. SPT results were far below the FIFO. Comparing results of the best model from the left side of Figure \ref{fig:test} with results for FIFO from the left side of Figure \ref{fig:test_comp}, we can see that cumulative reward for deep learning models is larger in the majority of episodes.

\begin{figure}
    \includegraphics[width=\textwidth]{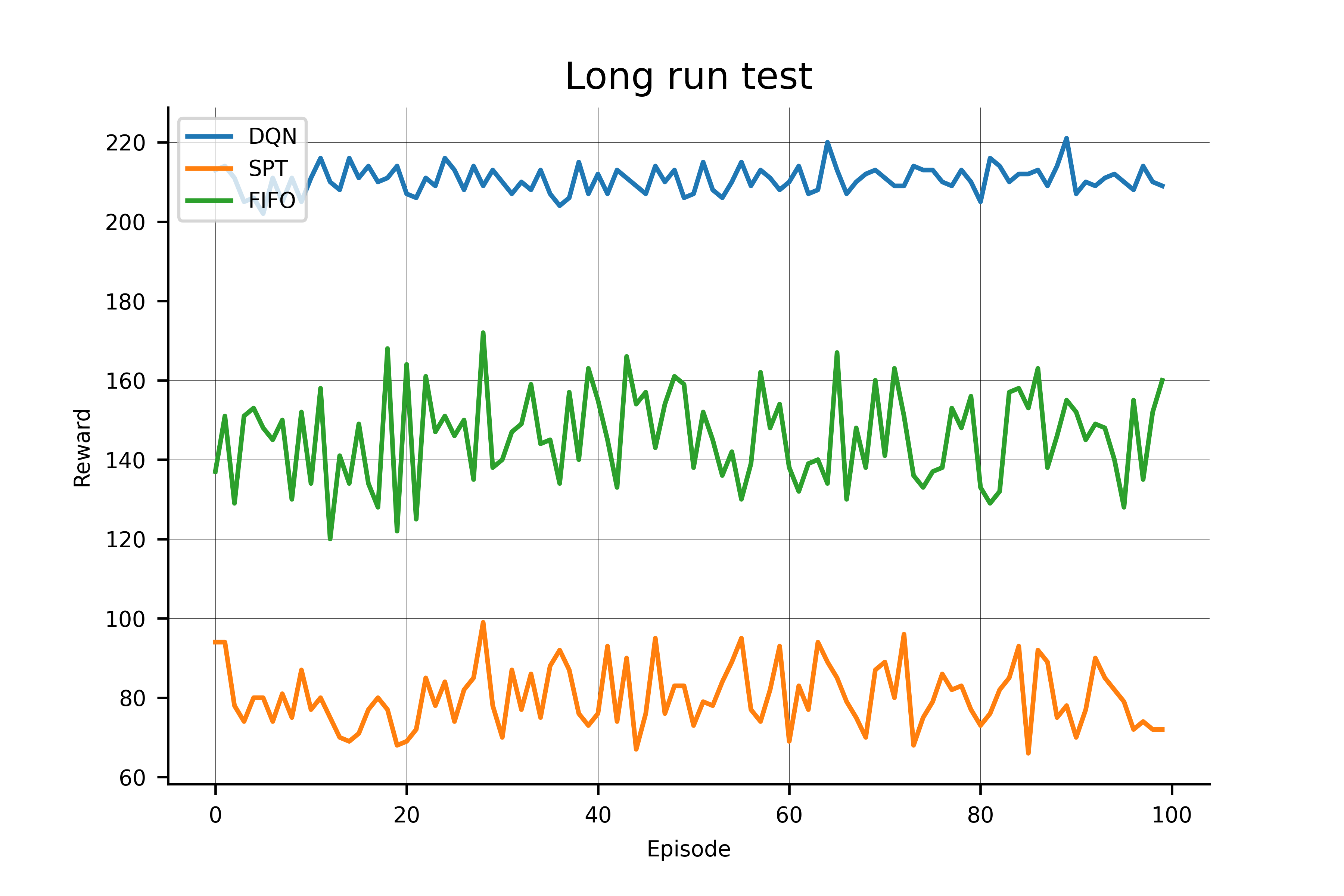}
    \caption{Long run test for best model achieved during training compared to FIFO and SPT approaches. Each episode lasted 5000 time-steps. }
    \label{fig:best}
\end{figure}

The improvement achieved by the deep RL model with each episode lasting $400$ steps is not large considering its absolute value. Median FIFO agent's reward oscillates around $11$, while median deep RL's around $14$. The question that arises here is whether this relation will hold with long (potentially infinitely) lasting episodes? To answer it we conducted an experiment with $100$ episodes with $5000$ steps each. The results are presented in Figure \ref{fig:best}. We can see that the gap between rewards for DQN model and for FIFO increased. Average episode reward for DQN was $210.52$, while for FIFO $145.84$ and $80.2$ for SPT.

\section{Conclusions and future work}
\label{sec:conc} 

In this paper we applied double deep reinforcement learning for the purpose of resource allocation in business processes. Our goal was to simultaneously optimize resource allocation for multiple processes and resources in the same way as it has to be done in the real-world scenarios. 

We proposed and implemented a dedicated simulation environment that enables agent to improve its policy in an iterative manner obtaining information about next states and rewards. Our environment is thus similar to OpenAI's Gym. We believe that along with processes' definitions it may serve as an universal testing suite improving reproducibility of the results for different resource allocation strategies.

We proposed a set of rules for defining business processes suites. They are the formal representation of real-world business process environments.  

The results of double DQN algorithm for resources allocation was compared with two strategies based on common heuristics: FIFO and SPT. Deep RL approach obtained results that are 44\% better than FIFO and $162\%$ better than SPT. We were not able to directly compare our results to previously published studies as they are relatively hard to reproduce. This was one of the main reasons to publish the code of both our simulation engine and training algorithm. We can see this as a first step towards a common platform that will allow different resource allocations methods to be reliably compared and assessed (in a similar manner to OpenAI Gym). 

As for future work, it would be very interesting to train a resource allocation agent for a business process suite with a larger number of business processes that would be more deterministic compared to those used in this study. Such a setup would put some light on a source of complexity in the training process. 

Number of potential actions and neural networks' outputs is a significant obstacle in applying proposed method for complex business process suites with many processes and resources. In our future work we plan to investigate other deep reinforcement learning approaches, such as proximal policy optimization, which tend to be more sample efficient than standard double DQN. 

\textbf{Reproducibility} The source code of Simulator Engine, deep RL model and test examples can be found on: https://github.com/kzbikowski/ProcessGym

\bibliographystyle{unsrt}
\bibliography{biblio}

\end{document}